\documentclass{bmvc2k}

\title{Local Feature Extraction from Salient Regions by Feature Map Transformation}
\addauthor{Yerim Jung}{annie_23@inha.edu}\\
\addauthor{Nur Suriza Syazwany}{surizasyazwany@inha.edu}\\
\addauthor{Sang-Chul Lee}{sclee@inha.ac.kr}\\

\addinstitution{
Department of Electrical\\
and Computer Engineering\\
Inha University\\
Incheon, South Korea
}

\usepackage{multirow}
\usepackage{multicol}
\usepackage{adjustbox}
\usepackage{amssymb}

\definecolor{ocre}{rgb}{0,0,.4}
\usepackage[font={color=ocre}]{caption}
\runninghead{JUNG ET AL.}{Local Feature Extraction from Salient Regions}

\begin{document}
\maketitle

\begin{abstract}
Local feature matching is essential for many applications, such as localization and 3D reconstruction. However, it is challenging to match feature points accurately in various camera viewpoints and illumination conditions. In this paper, we propose a framework that robustly extracts and describes salient local features regardless of changing light and viewpoints. The framework suppresses illumination variations and encourages structural information to ignore the noise from light and to focus on edges. We classify the elements in the feature covariance matrix, an implicit feature map information, into two components. Our model extracts feature points from salient regions leading to reduced incorrect matches. In our experiments, the proposed method achieved higher accuracy than the state-of-the-art methods in the public dataset, such as HPatches, Aachen Day-Night, and ETH, which especially show highly variant viewpoints and illumination.
\end{abstract}

%-------------------------------------------------------------------------
\section{Introduction}
\label{sec:intro}
Extracting and describing local features for matching is essential, especially in computer vision tasks that involve image matching, searching, tracking, and 3D reconstruction~\cite{heinly2015reconstructing,svarm2016city,noh2017large}. Feature matching focuses on three main phases when given two similar images are to be matched: feature detection, feature description, and feature matching~\cite{lowe2004distinctive,cheng2014fast}. The primary goal of feature matching is to optimize matching accuracy while minimizing the memory footprint of earlier applications. The extracted features should be sparse, highly repeatable, and precise. Each image's salient features, such as its corners, are initially recognized as interest points during the detection phase. Then, local descriptors are extracted based on the neighborhood regions of these interest points and used in the matching algorithms. 

Classical approaches~\cite{lowe2004distinctive,bay2006surf} concentrate on the \textit{detect-then-describe} method, where they first detect the points by analyzing the gradient of the image before describing the points with directional information. Furthermore, these approaches~\cite{lowe2004distinctive,bay2006surf} have shifted research trends by the emergence of deep learning methods~\cite{tian2017l2,zagoruyko2015learning,detone2018superpoint}. Since deep convolutional neural networks (DCNNs) can automatically learn features, mimic traditional detector behaviors, and process complex images, CNN methods have achieved remarkable performance than before~\cite{luo2020aslfeat,tian2020d2d,tyszkiewicz2020disk}. These data-driven methods concentrate on sparse points by leveraging descriptors' information for corresponding points. At the same time, several networks~\cite{detone2018superpoint,revaud2019r2d2} attempted to achieve better performance by influencing detection and description simultaneously, with improved repeatability and sparsity of detected points. Detector-free local feature matcher~\cite{sun2021loftr,zhou2021patch2pix} and a decoupled pipeline for a detection and description module were also studied~\cite{li2022decoupling}.

Despite these achievements, there is insufficient consideration for light and structure information in an image. Examining this information to robustly locate and define matched points, regardless of camera viewpoint or illumination variance, is critical. Nighttime images are challenging due to the uncertainties of light and structure~\cite{sun2021loftr}. When viewpoints change significantly, it is also difficult to match correctly~\cite{balntas2017hpatches}. Although some studies investigated this viewpoint and light information to apply in the local feature domain, they used only hand-crafted ways such as detecting corners or simply rotating features~\cite{pautrat2020online,liu2019gift,melekhov2020image}.

In this work, we propose a new strategy that uses both style and structure information to address the issue of mismatches in image variance. Specifically, we apply the concept of Instance Selective Whitening (ISW) loss, introduced by RobustNet~\cite{choi2021robustnet}, where the features are transformed with implicit information about the style component to have robustness under variations in light. Since there are limitations in this idea and considers only the style factor,  we revised ISW to consider the structure factor and apply it to the local feature field. Furthermore, we focus on salient points to reduce matching time.

\noindent\textbf{Contributions.} 
In this paper, we propose a framework that addresses the problems from a different light and structural information. We first extract features using Feature Map Generation (FMG) module. Then, Feature Map Transformation (FMT) module divides extracted features into two components: style and structure matrix. Each component independently learns the information gathered by the learned feature. Consequently, regardless of any changes in any component, the feature map will still select salient and matchable points. We introduce a loss function to maximize the influence of the structure information and minimize the style information in the feature map. The main contributions are summarized as follows:

 \begin{itemize}
%\vspace{-2mm}
\setlength\itemsep{-0.1em}
\item We overcome the limitation of feature matching in image variance by distinguishing between structure- and style-dependent features and transforming the feature maps. 
\item We propose Feature Map Transformation (FMT) module, exploiting an existing style transfer concept that concentrates only on style components, to transform the feature map while training to make it focus on salient features.
\item Extensive experiments on different benchmark datasets demonstrate that the proposed method can achieve high accuracy in matching tasks in a short time and with fewer parameters.
\end{itemize}

\section{Related Work}
\label{sec:related}
\noindent\textbf{Local feature learning.} 
The joint learning of feature detectors and descriptors requires a unified network to construct feature maps and allows the two tasks to share the majority of computations for improved performance. DELF~\cite{noh2017large} proposed an image retrieval technique that learns local features as a by-product of a classification loss combined with an attention mechanism to improve performance on large-scale images. It outperforms images under changing light conditions but has limitations in terms of structural variation. SuperPoint~\cite{detone2018superpoint} suggested a method for learning from the manual annotation of significant points on simple images such as corners and edges. However, because of their low repeatability and descriptor accuracy, it has many outliers, so the matched points tend to be mistakenly judged. R2D2~\cite{revaud2019r2d2} has overcome this issue by learning the descriptor reliability in parallel with the detection and description phases and only selecting both repeatable and reliable keypoints with respect to the descriptor. 

 To find only the matchable points, D2-Net~\cite{dusmanu2019d2} proposed a describe-and-detect method for joint detection and description that uses a single CNN with shared weights. The detection is based on the entire channel's local maxima and the shape map's spatial dimensions. DISK~\cite{tyszkiewicz2020disk} applied reinforcement learning on an end-to-end network inspired by D2-Net~\cite{dusmanu2019d2} that relied on policy gradients. Furthermore, ASLFeat~\cite{luo2020aslfeat} demonstrated significant improvement using a score map that used local shape estimation to select matching points.

\noindent \textbf{Feature covariance.}
Previous research~\cite{gatys2015texture,gatys2016image} proposed that image style information is considered via feature correlations such as a gram or a covariance matrix. Since then, feature correlation has been applied to several different research areas, including style transfer~\cite{li2017universal}, image-to-image translation~\cite{cho2019image}, domain adaptation~\cite{roy2019unsupervised,sun2016deep}, and network architecture~\cite{luo2017learning,pan2019switchable,huang2018decorrelated}. Whitening transformation (WT)~\cite{li2017universal,cho2019image,pan2019switchable}, which eliminates feature correlation and assigns unit variance to each feature, aids in the removal of style information from the feature representations.

Since region-specific styles and region-invariant content are simultaneously written to the covariance vector of the feature maps, whitening all the correlation components reduce feature identification and distort the boundaries of objects~\cite{li2017universal,li2018closed}. RobustNet~\cite{choi2021robustnet} proposed the ISW loss, which extracted only the style information to solve the problem. We want to focus on style and structure information, so we modify the ISW loss to satisfy our objective.

\section{Method}
\subsection{Feature Map Generation Module}
\begin{figure}[t]
\centering
\includegraphics[width=1\linewidth]{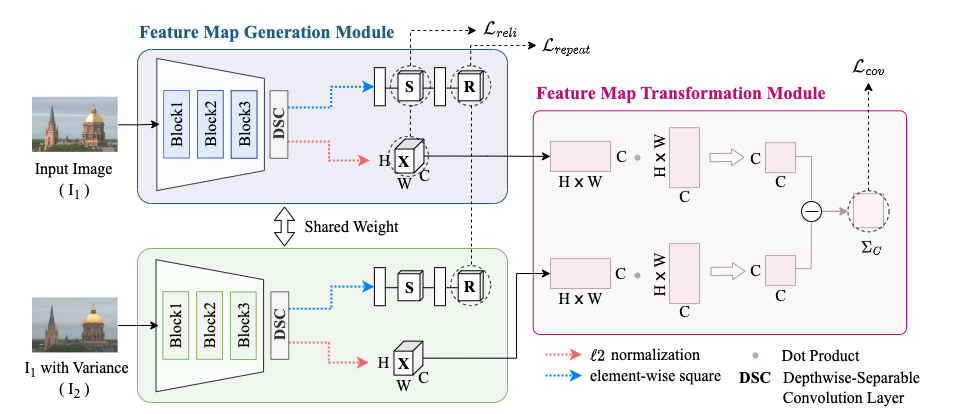}
\caption{\textbf{Proposed framework.} Network consisting of a feature map generation module and a transformation module is shown. The reliability map $\mathbf{S}$ and repeatability map $\mathbf{R}$ learn the regions of points of interest, while the covariance matrix derived from the descriptor map $\mathbf{X}$ is used to transform the feature map to focus on saliency with style and structure information.}
\label{fig:main}
\end{figure}

Feature Map Generation (FMG) module first extracts the features of an image pair, $I_{1}$ and $I_{2}$ independently, which outputs two branches: descriptors and point extraction feature maps. The point extraction branch consists of two feature maps. One produces another with a 1$\times$1 convolution layer; the former is a reliability map $\mathbf{S}$ and the latter in repeatability map $\mathbf{R}$. The covariance matrix derived from the descriptor map $\mathbf{X}$ is used to transform the feature map to focus on saliency with style and structure information. Then FMG module uses feature maps $\mathbf{X}$, $\mathbf{S}$, and $\mathbf{R}$ to calculate the loss functions for repeatability and reliability. The FMG module's network architecture differs from R2D2~\cite{revaud2019r2d2} but uses the same loss functions in this part. So only the architecture part will be described. The proposed method pipeline is shown in Figure~\ref{fig:main}.

As introduced in MobileNet~\cite{howard2017mobilenets}, depthwise separable convolution (DSC) is a factorization convolution method that significantly reduces computation and model size with new representations. Motivated by this, we used the DSC to focus on the salient area. Inspired by ~\cite{revaud2019r2d2,luo2020aslfeat}, we adopt a modified L2Net~\cite{tian2017l2}—where the last layer is replaced by a set of three consecutive layers—in our backbone network to extract feature information from an image pair. Since the input images are image pairs, two backbone networks are needed. We use weight sharing in the DSC layer when adding it behind the backbone network to reduce the model weight. Furthermore, the relationship between descriptors and points of interest can be maintained by sharing the weights. The backbone network then generates three feature maps, $\mathbf{X}$ by $\ell$2 normalization, $\mathbf{S}$ by the element-wise square, and $\mathbf{R}$ obtained from $\mathbf{S}$ with a 1$\times$1 convolution layer. In contrast to ~\cite{revaud2019r2d2}, $\mathbf{S}$ and $\mathbf{R}$ come from the same branch since they depend on one another. We assumed that $\mathbf{S}$ is a feature map that learns a point that reduces the matching distance, which might affect in high repeatability rate in $\mathbf{R}$. Therefore, the model weight gets lighter and develops more robust features through information sharing.

FMG module calculates the reliability loss, $\mathcal{L}_{reli}$, to get discriminative feature points. Let $\mathbf{X}_{ij}$ be the local descriptor in each pixel $(i, j)$ of the image $I_{1}$; we then predict the individual reliability scores $\mathbf{S}_{ij}$ from $\mathbf{X}_{ij}$ and $\mathbf{X}^{'}_{uv}$. Here, we specify the exact coordinate $(u, v)$ that corresponds to $(i, j)$, knowing the ground truth correspondence mapping $T$, where $T\in \mathbb{R}^{H\times W\times 2}$ is the ground truth correspondence between image $I_{1}$ and $I_{2}$. $\mathbf{X}_{ij}$ is compared with $\mathbf{X}^{'}_{uv}$, where $\mathbf{X}^{'}_{uv}$ is extracted from $I_{2}$. Then, average precision is used to calculate $\mathcal{L}_{reli}$, optimized with a differentiable approximation~\cite{he2018local,revaud2019r2d2}, using $\mathbf{S}_{ij}$.

In addition, FMG module calculates the repeatability loss, $\mathcal{L}_{repeat}$, for extracting repeatable feature points as in~\cite{revaud2019r2d2}. It uses peakiness prediction and similarity between feature pairs from input pair images. For similarity, Let $\mathbf{R}$ and $\mathbf{R^{'}}$ be the repeatability maps corresponding to $I_{1}$ and $I_{2}$. We set $\mathbf{R}^{'}_{T}$ to be a map in which $\mathbf{R^{'}}$ is transformed by the ground truth homography relationship between image pairs $I_{1}$ and $I_{2}$. Because the prime objective is to predict keypoints with high repeatability, we train the network so that the positions of the local maxima in $\mathbf{R}$ are covariant to the actual picture transformations, such as viewpoint structures or light shifts. Assuming that all the local maxima of $\mathbf{R}$ coincide with the local maxima of $\mathbf{R}^{'}_{T}$, we define the loss function. The basic concept is to maximize the cosine similarity between $\mathbf{R}$ and $\mathbf{R}^{'}_{T}$ such that the two heatmaps are identical and their maxima identically coincide. The loss may remain at a particular constant that may terminate the learning process, so we prevent this using the peakiness prediction. The final repeatability loss is calculated by considering both similarity and the peakiness of the input image pair.

\subsection{Feature Map Transformation Module}

One problem with feature matching is structural perspective and lighting (style) differences. The style corresponds to noise, such as weather and light. Concentrating on the structure, which relates to the image's point of view or the edges, can result in better results. Feature Map Transformation (FMT) module suppresses the style information by performing task adaptation and supplementing the existing WT loss with an extensive transformation. We improve the previous loss by including structural information to overcome the limitations of using only existing styles.

\noindent\textbf{Style/Structure Covariance Matrix.}
Previous studies~\cite{li2019episodic,choi2021robustnet} claimed that applying WT to each instance of style transfer could successfully erase style information. WT is a linear transformation that equalizes the variance term in each channel to one and reduces the covariances between channels to zero. The intermediate feature map is $\textbf{X}\in\mathbb{R}^{C\times HW}$, where $C$ is the number of channels, and $H$ and $W$ are the height and width of the feature map, respectively. The covariances between pairs of channels can be defined as follows:

 \begin{equation}
\Sigma =\frac{1}{HW}\left (  \textbf{X}-\mathbf{\mu} \cdot \textbf{O}^{\top }\right )\left (  \textbf{X}-\mathbf{\mu} \cdot \textbf{O}^{\top }\right )^{\top }\in \mathbb{R}^{C\times C}
  \label{eq:two}
\end{equation}

\noindent where $\textbf{O}\in\mathbb{R}^{HW}$ is a column vector of ones, and the $\mathbf{\mu}$ and $\Sigma$ are the mean vector and covariance matrix, respectively. When the loss function is designed so that the elements of the covariance matrix $\Sigma$ decrease, the feature extraction is less affected by the style element in extracting features from the input image. This is because the feature map generation lacks the information of style elements. Based on this concept, we adopt a method of transforming feature maps using style elements. Furthermore, we use information about the structure, the leftover elements in the gram matrix beside style elements, to design the loss function in the direction of expansion rather than suppression.

\begin{figure}[t]
\centering
\includegraphics[width=1\linewidth]{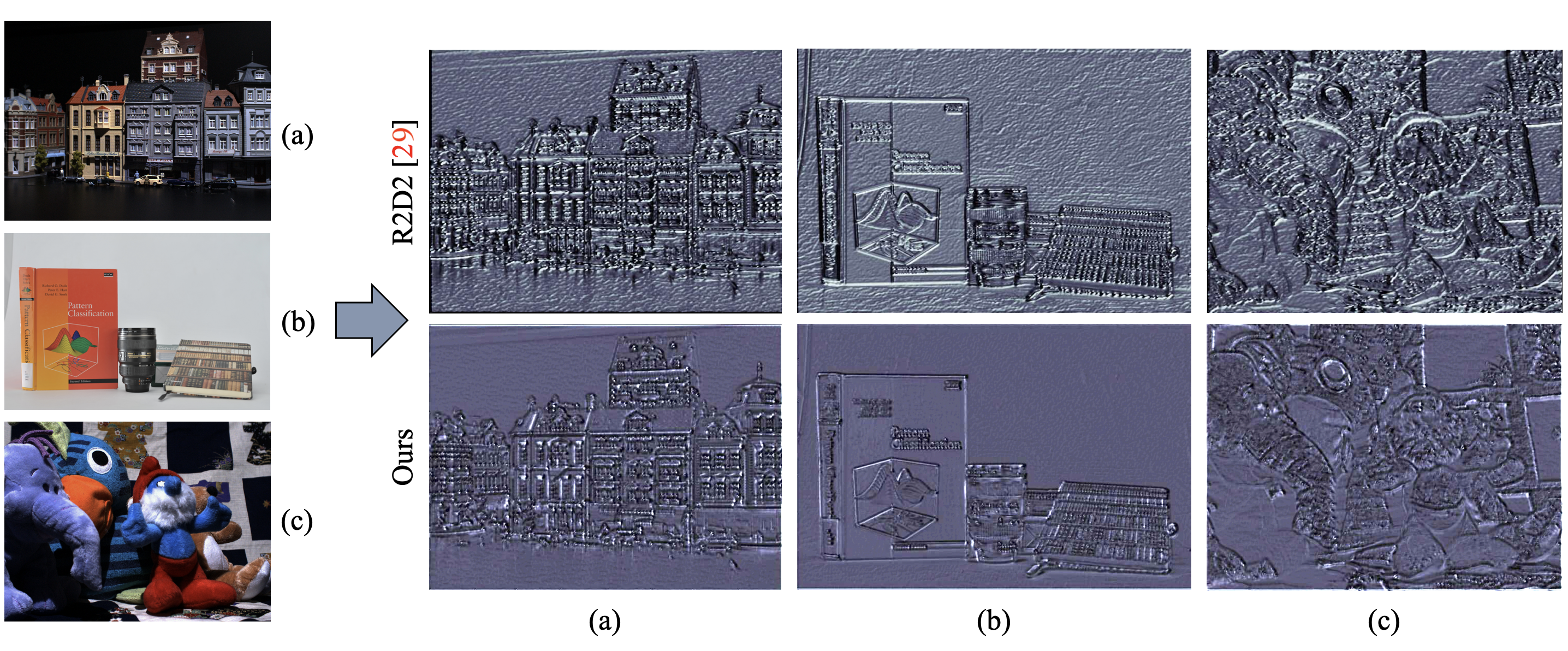}
\caption{\textbf{Transformed feature map.} Comparison of feature maps from each network indicates the tendency in which we change the feature map; eliminates style information refers to noise, while structure complexity enlarges.}
\label{fig:cov}
\end{figure} 

\noindent\textbf{Transformation Loss.} In order to transform and manipulate feature maps using the aforementioned style and structure information, we introduce the FMT module which use feature map $\textbf{X}$ to calculate $\mathcal{L}_{cov}$. This is done since the descriptor has the most information among the three output feature maps from the FMG module. FMT module then separates style and structure characteristics from the feature representation's higher-order statistics by selectively modifying each attribute differently. Replacing the old feature map $\textbf{X}$ with a standardized feature map $\mathbf{X}_{s}$ simplifies the optimization process of both the diagonal and off-diagonal elements of the covariance matrix simultaneously~\cite{ulyanov2016instance,choi2021robustnet}. $\Sigma_{s}$ is a matrix made from the standardized feature map $\mathbf{X}_{s}$, and we define $\Sigma_{s}$ as follows:

\begin{equation}
\Sigma_{s}=\frac{1}{HW} \mathbf{X}_{s}\cdot \mathbf{X}_{s}^{\top }\in \mathbb{R}^{C\times C}
  \label{eq:three}
\end{equation}

\noindent After obtaining the covariance matrices $\Sigma_{s}(I_{1})$ and $\Sigma_{s}(I_{2})$, we calculate the difference between the two matrices to produce matrix $\Sigma_{C}$, as defined in Eq.~\ref{eq:five2}. Matrix $\Sigma_{C}$ indicates the sensitivity of the corresponding covariance to the photometric transformation~\cite{choi2021robustnet}. Elements with a high variance value retain the style information, whereas elements with a low variance value retain the structure information. We use absolute value notation with vertical bars.

\begin{equation}
\Sigma_{C}=\left|\Sigma_{s}(I_{1})-\Sigma_{s}(I_{2}) \right|
   \label{eq:five2}
\end{equation}

We cluster the style and structural components in equal amounts, using the mean value to determine the threshold for separating the two components. If a matrix $\Sigma_{C}$ element is greater than the threshold, that element is classified as a style factor, while the rest of the elements are classified as structural factors. This definition is established because the prominent factors of the matrix $\Sigma_{C}$ are assumed to imply the style factor~\cite{choi2021robustnet}. In this case, the style factor refers to changes in light or color, and the structural factor refers to complexities with many objects, edges, or viewpoints. The proposed loss function, $\mathcal{L}_{cov}$ is formulated as follows:

\begin{equation}
\mathcal{L}_{cov}=\begin{cases}
 \mathbb{E}\left [ \left \| \Sigma_{C} \odot \mathbf{M}_{sty} \right \|_{1} \right ]& \text{ if } \Sigma_{C} >  \mathbf{\mu} \\ 
\ 1-\mathbb{E}\left [ \left \| \Sigma_{C} \odot \mathbf{M}_{str} \right \|_{1} \right ] & \text{ otherwise}
\end{cases}
  \label{eq:six}
\end{equation}

\noindent where $\mathbb{E}$ and $\mathbf{M}\in\mathbb{R}^{C\times C}$ are the arithmetic mean and a mask matrix. $\mathbf{M}_{sty}$ and $\mathbf{M}_{str}$ are masks that select style and structure values, and $\odot$ is element-wise multiplication. Finally, the total loss function can be represented by Eq.~\ref{eq:twelve}, with each weight $\lambda_{i}$ are empirically tuned to the optimal ratio of 1:1:2. We define $i\in \left\{ 1, 2, 3\right\}$ because the number of loss terms is three. We strengthen the argument that adding the transformation loss is superior in selecting only salient features when comparing the feature map of R2D2~\cite{revaud2019r2d2} with ours in Figure~\ref{fig:cov}.

\begin{equation}
\mathcal{L}_{total}=\lambda_{1}\cdot \mathcal{L}_{reli}+\lambda_{2}\cdot \mathcal{L}_{repeat}+\lambda_{3}\cdot \mathcal{L}_{cov}
  \label{eq:twelve}
\end{equation}

This aggregated loss function is used to select salient points to minimize the prediction of the less informative regions, such as the sky or ground. FMT is inducing the feature map transformation so that the proposed transformation loss function could extract robust features if the location where the image is taken is in the same place, regardless of the change in structure and style.

\section{Experiment}

\subsection{Implementation details}
\textbf{Training.}
We apply Adam to optimize the network for 25 epochs with a fixed learning rate of 0.0001, a weight decay of 0.0005, and a batch size of 8 pairs of cropped images of 192 by 192 pixels, as in R2D2~\cite{revaud2019r2d2}. Our experiment used the training dataset and ground-truth correspondences used in R2D2. Since our model uses the modified version of R2D2, we trained the network from scratch. Nonetheless, we fixed the patch size N used in the repeatability loss to 16 in all training parts to improve the performance of the transformation loss.

\noindent\textbf{Testing.}
We used the sum of different scales of images to diversify the resolution of the feature maps at test time. The descriptors were interpolated at the modified locations. This multi-scale feature extraction enables the extraction of more tentative keypoints and provides improved localization.

\noindent\textbf{Experiment Settings.}
This study used an NVIDIA GeForce RTX 3090 GPU and CUDA toolkit, version 11.2, with Python 3 and PyTorch 1.8 in the training environment.

\renewcommand{\arraystretch}{1}
\begin{figure}[t!]
    \begin{minipage}{0.60\linewidth}\hspace{5mm}\centering
		\includegraphics[width=1.05\textwidth]{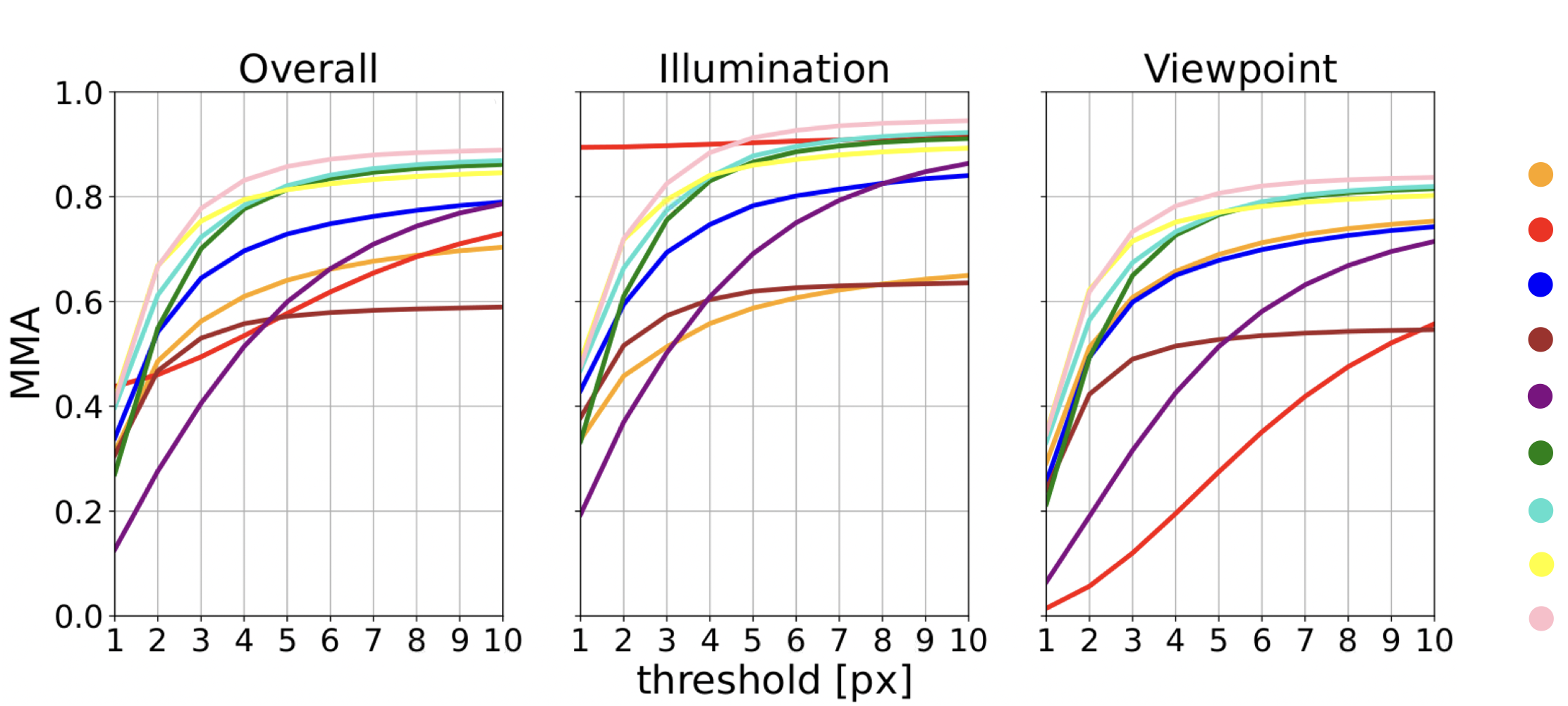}
		 \caption{\textbf{Quantitative Results on HPatches.} Comparison in terms of MMA on the HPatches dataset.}
		\label{figu:re}
	\end{minipage}
    \hfill
     \begin{minipage}{0.38\hsize}\hspace{5mm}\centering
    \begin{adjustbox}{width=1\columnwidth,center}
       \begin{tabular}{c||ccc}
\noalign{\smallskip}\noalign{\smallskip}\hline\hline
% \multirow{0}{0} 
&\multicolumn{3}{c}{MMA@3} \\
\cline{2-4}
&Overall&Illumi.&Viewp.\\ \hline
Hes.Aff.~\cite{perd2009efficient} & 56.24 &51.35 &60.79\\
DELF(new)~\cite{noh2017large} & 49.43 &\textbf{ 89.73} & 12.02 \\
SuperPoint~\cite{detone2018superpoint} &64.45 &69.38 &59.88 \\
LF-Net~\cite{ono2018lf} & 53.01 &57.31 &49.02 \\
D2-Net~\cite{dusmanu2019d2} &39.76 &44.99 &34.91 \\
R2D2~\cite{revaud2019r2d2} &70.06 &75.56 &64.96\\
ASLFeat~\cite{luo2020aslfeat} & 72.28& 75.47 &68.28\\
DISK~\cite{tyszkiewicz2020disk}  & 75.34 & 79.43 &71.53\\
Ours  &\textbf{78.41} &\textit{83.22} &\textbf{73.94}\\
\hline
\hline
\end{tabular}
\end{adjustbox}
\vspace*{5pt}
 \captionof{table}{\textbf{MMA@3 on HPatches.} Comparison at 3px threshold.}
\label{tab:singlebesti}
\end{minipage}
\end{figure}

\subsection{Feature Matching}
\noindent\textbf{Quantitative Evaluation.}
We evaluated the performance of selecting meaningful points by calculating mean matching accuracy (MMA). If the distance between the converted point and the reference point exists within the threshold based on the ground truth homography metric, the converted point is classified as a correct conversion point. Figure~\ref{figu:re} illustrates the comparisons on the H-Patches dataset~\cite{balntas2017hpatches}, with MMA measured at various error thresholds. Figure~\ref{figu:re} was drawn from the cache data provided by the D2-Net~\cite{dusmanu2019d2} repository. The comparison was performed with DELF~\cite{noh2017large}, SuperPoint~\cite{detone2018superpoint}, LF-Net~\cite{ono2018lf} mono and multi-scale D2-Net~\cite{dusmanu2019d2}, R2D2~\cite{revaud2019r2d2}, ASLFeat~\cite{luo2020aslfeat}, DISK~\cite{tyszkiewicz2020disk}, and a hand-crafted Hessian affine detector with a
RootSIFT descriptor~\cite{perd2009efficient}. Our network denoted as "Ours" outperformed almost all state-of-the-art networks. Regarding illumination, DELF~\cite{noh2017large} outperformed our method since it identifies key points in a low-resolution feature map with a fixed grid. However, ours still exhibited the highest performance in terms of the five-pixel threshold because of the fixed grid of keypoints without spatial variation in this subgroup. The overall scores for the illumination dataset and viewpoints and their respective individual scores are presented in Table~\ref{tab:singlebesti}, with the MMA threshold set to 3.

\begin{figure}
\centering
\includegraphics[width=1\linewidth]{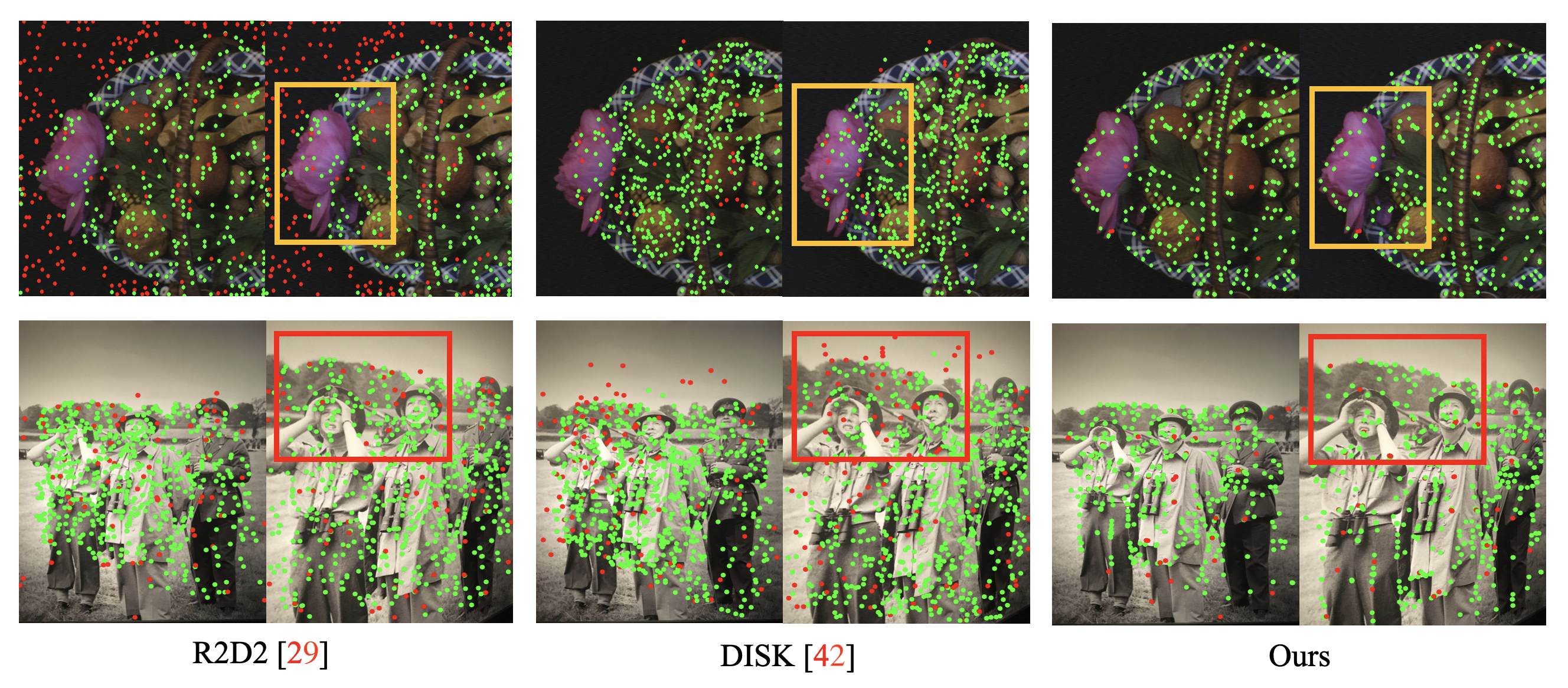}
\caption{\textbf{Qualitative Results on HPatches.} There are illustrated pairs of illumination and viewpoint variations. Green dots represent points that are correctly matched, whereas red dots represent points that are incorrectly matched. The boxed area provides significantly better outcomes than previous methods by focusing on the salient region.}
\label{fig:detail2}
\end{figure}

\noindent\textbf{Qualitative Evaluation.}
Figure~\ref{fig:detail2} shows the comparison between our baseline network R2D2~\cite{revaud2019r2d2} and the current state-of-the-art method DISK~\cite{tyszkiewicz2020disk} with a severe change in illumination and viewpoint in the first and second row, respectively. The experiment was done with the nearest neighborhood matching with 3 points of error threshold. When we look at the yellow box, we see that we succeeded in focusing on structural information. The structurally less focused network failed to match in this part. In the red box, it can be seen that the error rate is lowered by focusing less on the background or natural objects corresponding to noise. The matching time is also shorter, shown in Figure~\ref{tab:mat}.

\subsection{Visual Localization} 
In a local reconstruction job~\cite{sattler2017large}, we evaluate our technique on the Aachen Day-Night dataset v1.1~\cite{sattler2018benchmarking} as in D2-Net~\cite{dusmanu2019d2}. In this section, we present the results of an optical localization task. Given the daytime photographs with known camera positions, our goal is to identify the nighttime image of the same area. The known location of the daytime photos in each set is used to triangulate the 3D structure of the scene after extensive feature matching. Finally, these 3D models are used to locate the query photographs taken at night. We followed the guidelines for Visual Localization Benchmark, for which we used our matches as input for a pre-defined visual localization pipeline based on COLMAP~\cite{schonberger2016structure,schonberger2016pixelwise}. We also adopted hierarchical localization~\cite{sarlin2019coarse} in every network for higher performance. This pipeline was then used to build an SfM model with the registered test photos. We used NetVlad~\cite{arandjelovic2016netvlad} for the global feature and the nearest neighborhood for matching. The percentages of properly localized photos under three error levels are reported in Table~\ref{tab:3}. Our results demonstrate our method’s strong generalization capabilities because of its high localization performance compared with SuperPoint~\cite{detone2018superpoint}, D2-Net~\cite{dusmanu2019d2}, R2D2~\cite{revaud2019r2d2} and DISK~\cite{tyszkiewicz2020disk}. Our network performed fairly well when compared with matcher methods~\cite{sun2021loftr,zhou2021patch2pix}. 

 \renewcommand{\arraystretch}{1}
 \begin{table}[t!]
    \begin{minipage}{0.66\linewidth}\hspace{5mm}\centering
\begin{adjustbox}{width=1\columnwidth,center}
    \begin{tabular}{c||ccc|ccc}
    \noalign{\smallskip}\noalign{\smallskip}\hline\hline
     \multirow{2}{*}{} & \multicolumn{3}{c|}{Day} & \multicolumn{3}{c}{Night} \\
    \cline{2-7}
          & 0.5m, $2^{\circ}$ & 1m, $5^{\circ}$ & 5m, $10^{\circ}$ & 0.5m, $2^{\circ}$  & 1m, $5^{\circ}$ & 5m, $10^{\circ}$ \\
    \hline
     SuperPoint~\cite{detone2018superpoint} & 85.3 & 91.9 & 94.5&	58.6 & 74.3 & 85.9\\
     D2-Net~\cite{dusmanu2019d2} &81.6 & 89.3 & 96.2	&62.8 &80.6 & 92.7 \\
     R2D2~\cite{revaud2019r2d2} &89.9 & 95.4 &  98.4& 	69.6 &  85.9 & 96.3  \\
     DISK~\cite{tyszkiewicz2020disk} &- & - &  -&  \textbf{72.3}& 86.4&\textbf{97.9}\\
      Ours &\textbf{90.4} & \textbf{96.1}& \textbf{98.9}&	\textbf{72.3} &\textbf{89.0} & \textit{96.9}\\
     \hline\hline
     LoFTR$^{*}$~\cite{sun2021loftr} &- & - &  -& \underline{72.8} & 88.5 & \underline{99.0} \\
    Patch2Pix$^{*}$~\cite{zhou2021patch2pix} &86.4&93.0&97.5&	\underline{72.3}&88.5&\underline{97.9}\\
    \hline\hline
    \end{tabular}
    \end{adjustbox}
    \vspace*{2pt}
    \caption{\textbf{Aachen Evaluation.} Comparison on Aachen for the visual localization task. * are marked for matcher methods.}
    \label{tab:3}
	\end{minipage}
    \hfill
     \begin{minipage}{0.33\hsize}\centering
     \includegraphics[width=0.92\textwidth]{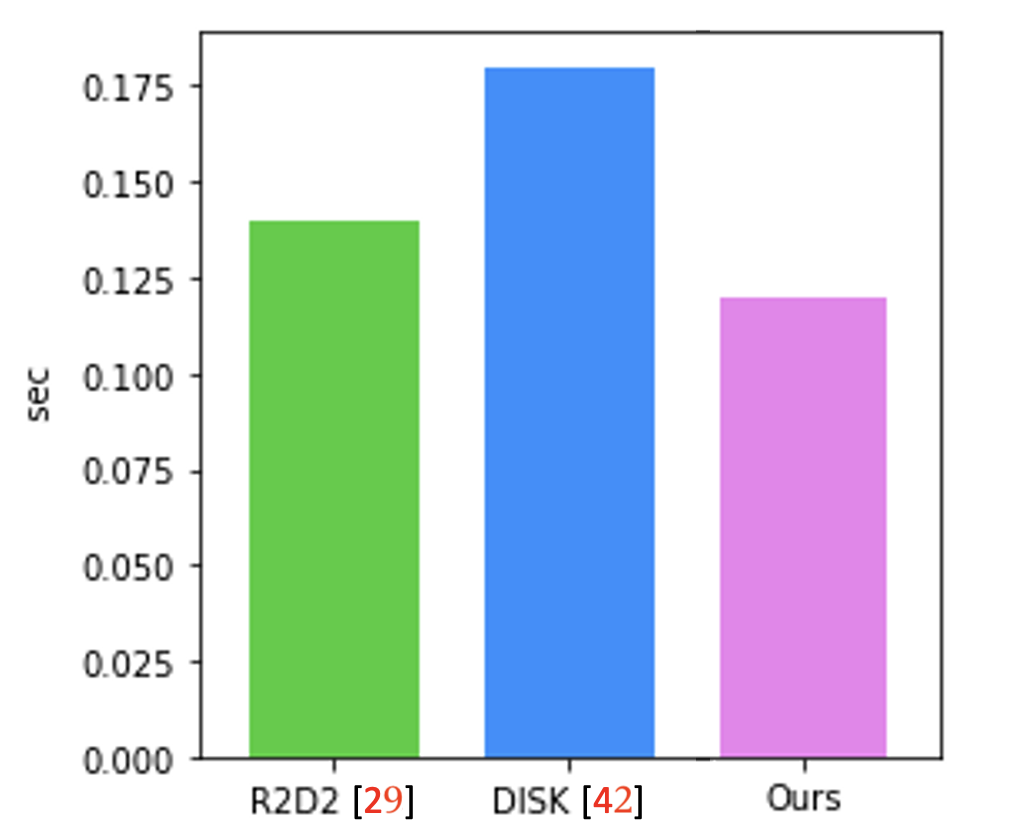}
		 \captionof{figure}{\textbf{Matching Time.} Comparison of matching time.}
		\label{tab:mat}
\end{minipage}
\end{table}

\subsection{3D Reconstruction} 
For 3D reconstruction evaluation, we used ETH-Microsoft Dataset~\cite{schonberger2017comparative} and for the evaluation protocols, we ran SfM algorithm by COLMAP~\cite{schonberger2016structure,schonberger2016pixelwise}. In Table~\ref{tab:eth}, we compare our method with the results presented in~\cite{luo2020aslfeat}, but only the jointly learned models. Compared with other methods, since the number of matched points was less than 100K, we do not provide the comparison with R2D2~\cite{revaud2019r2d2}. For sparse reconstruction, we report the number of registered images ($\#$ Reg), the number of sparse points ($\#$ Sparse), tracked length (Track), and reprojection error (Reproj). Table~\ref{tab:eth} data shows that our network performs favorably against previous methods on the 3D reconstruction task. Furthermore, the number of registered images obtained the best value, and the number of sparse points also had the best or second-order value. It can be interpreted that the selection of salient points was made well.

\renewcommand{\arraystretch}{1.2}
\begin{table}
\centering
\begin{adjustbox}{width=0.98\columnwidth,center}
\begin{tabular}{c||cccc|cccc|cccc}
\hline\hline
\multirow{3}{*}{} 
& \multicolumn{4}{c|}{Madrid Metropolis 1344 images}
&\multicolumn{4}{c|}{Gendarmenmarkt 1463 images}
&\multicolumn{4}{c}{Tower of London 1576 images} 
\\
 
 \cline{2-13} &$\#$ Reg& $\#$ Sparse& Track &Reproj&$\#$ Reg& $\#$ Sparse& Track &Reproj&$\#$ Reg& $\#$ Sparse& Track &Reproj\\
 \hline
 SuperPoint~\cite{detone2018superpoint} &438 &29K &9.03& 1.02px  &967& 93K& 7.22 & \textbf{1.03px}&681& 52K& 8.67 &0.96px  \\
D2-Net~\cite{dusmanu2019d2} &495& \textbf{144K}& 6.39& 1.35px  & 965& 310K& 5.55& 1.28px & 708& 287K &5.20& 1.34px \\
ASLFeat~\cite{luo2020aslfeat} & 649& 129K& \textbf{9.56}& \textbf{0.95px}&1061& 320K& \textbf{8.98}& 1.05px  &846& 252K &\textbf{13.16 }&\textbf{0.95px} \\
Ours & \textbf{766} & 142K & 8.13 & 1.19px & \textbf{1316} &  \textbf{516K} & 6.81 & 1.19px & \textbf{1186} & \textbf{315K} & 8.63 &1.21px\\

% \noalign{\smallskip}\noalign{\smallskip}\hline\hline
% \multicolumn{2}{c|}{} &$\#$Reg.& $\#$ Sparse.& Track &Reproj.\\
% \hline
% \multirow{5}{*}{\parbox{1.8cm}{Madrid Metropolis 1344 images}}
% % & RootSIFT &500& 116K &6.32 &\textbf{0.60px} \\
% % & GeoDesc &495 &\textbf{144K}& 5.97 &0.65px \\
% & SuperPoint &438 &29K &9.03& 1.02px  \\
% & D2-Net (MS) &495& \textbf{144K}& 6.39& 1.35px   \\
% & ASLFeat (MS) & 649& 129K& \textbf{9.56}& \textbf{0.95px} \\
% & Ours & \textbf{766} & 142K & 8.13 & 1.19px  \\
% \hline
% \multirow{5}{*}{\parbox{1.8cm}{Gendar-menmarkt\\1463 images}} 
% % & RootSIFT & 1035& 338K& 5.52 &\textbf{0.69px}\\
% % & GeoDesc &1004 &\textbf{441K }&5.14& 0.73px\\
% & SuperPoint &967& 93K& 7.22 & \textbf{1.03px}  \\
% & D2-Net (MS) & 965& 310K& 5.55& 1.28px  \\
% & ASLFeat (MS) &1061& 320K& \textbf{8.98}& 1.05px \\
% & Ours & \textbf{1316} &  \textbf{516K} & 6.81 & 1.19px\\
% \hline
% \multirow{5}{*}{\parbox{1.8cm}{Tower of\\London\\1576 images}}
% % & RootSIFT & 804& 239K& 7.76& \textbf{0.61px} \\
% % & GeoDesc & 776& \textbf{341K}&6.71& 0.63px  \\
% & SuperPoint &681& 52K& 8.67 &0.96px \\
% & D2-Net (MS) & 708& 287K &5.20& 1.34px \\
% & ASLFeat (MS) &846& 252K &\textbf{13.16 }&\textbf{0.95px} \\
% & Ours & \textbf{1186} & \textbf{315K} & 8.63 &1.21px\\
\hline
\hline
\end{tabular}
\end{adjustbox}
\caption{\textbf{ETH Evaluation.} 3D reconstruction held with  ETH-Microsoft Dataset.}
\label{tab:eth}
\end{table}

%------------------------
\renewcommand{\arraystretch}{1}
\begin{table}[t]

% \begin{minipage}{.45\linewidth}
    \centering
    \begin{adjustbox}{width=0.98\columnwidth}
    \begin{tabular}{c||ccc|ccc|ccc}
    \hline\hline
    \multirow{2}{*}{} & \multicolumn{3}{c|}{HPatches} &\multicolumn{3}{c|}{Aachen (Day)} 
&\multicolumn{3}{c}{Aachen (Night)} 
\\
 
 \cline{2-10} 
&Overall&Illum&Viewp
& 0.5m, 2$^{\circ}$ & 1m, 5$^{\circ}$ & 5m, 10$^{\circ}$ 
& 0.5m, 2$^{\circ}$ & 1m, 5$^{\circ}$ & 5m, 10$^{\circ}$ \\
  \hline
     w/o $\mathcal{L}_{sty} \& \mathcal{L}_{str}$ 
     &70.06 &75.56 &64.96
    %  아래에 w/o style&structure 넣기
     &89.9 & 95.4 &  98.4& 	69.6 &  85.9 & 96.3       % <- Aachen Night
     \\
     w/o $\mathcal{L}_{sty}$ 
     & 72.08 & 78.04 & 66.55
    %  아래에 w/o style 넣기
     &89.8 & \textbf{96.1} & 98.7&	\textbf{73.3} & 88.5 &95.3      % <- Aachen Night
     \\
     w/o $\mathcal{L}_{str}$ 
     & 76.53 & 81.43 & 71.99
    %  아래에 w/o structure 넣기
    &89.7 & 95.8 & 98.5
    &69.6 & 84.8 &96.3      % <- Aachen Night
     \\
 w/o DSC
     &76.89 &81.95 &72.19
    %  아래에 w/ style&structure  넣기
      &89.9 & 95.3 & 98.2&	69.1 &85.9 & 93.7    % <- Aachen Night
     \\
     \hline
      w/ All
     &\textbf{78.41} &\textbf{83.22} &\textbf{73.94}
    %  아래에 w/ style&structure  넣기
     &\textbf{90.4} & \textbf{96.1}& \textbf{98.9}&	
    \textit{72.3} &\textbf{89.0} & \textbf{96.9}     % <- Aachen Night
     \\
    \hline\hline
    \end{tabular}
    \end{adjustbox}
    
% \end{minipage}
\vspace*{5pt}
\caption{\textbf{Ablation Studies.} Ablation experiment on MMA@3 and visual localization to see how each component affect transformation loss. We show that the best results occur when style and structure loss are included with depth-wise convolution.}
\label{tab:a2}
\end{table}

\subsection{Ablation Studies}

We validated the significance of the components that comprise our suggested transformation loss function by performing two ablation studies. We studied the presence of style and structural component losses to determine how each characteristic contributes to learning. $\mathcal{L}_{sty}$ and $\mathcal{L}_{str}$ denote loss function that use same subscript in the masks of Eq.~\ref{eq:six}. The result in Table~\ref{tab:a2} reveals that the matching accuracy is not only influenced by the style factor but also by the structure factor. This experiment confirms the efficiency of ISW when applied to our method and its ability to provide additional information about the structure. Furthermore, ablation studies were conducted with and without DSC layer. The performance improved using DSC layer, with a model weight reduction of 2 MB.

\section{Conclusion}
We proposed a robust network using self-transformation loss, which transforms a feature map that contributes to the repeatability of local features. We separated structure and style characteristics by clustering the covariance vector and influenced the feature of each characteristic. The feature maps were unified to make two feature maps closer by reducing the light component and sharpening the structure component. Consequently, this step increases the repeatability of the points, reduces outliers, and assigns robustly matched descriptors. A comparison with similar research reveals that our method more effectively extracts robust matching points in various scenes. Nevertheless, the current work has limitations. Points tend to be retrieved by clustering them in a particular region. This analysis reveals that adaptively selecting from sparse points is a promising avenue for future research.

\section{Acknowledgment}
This work was supported in part by the National Research Foundation of Korea (NRF) under Grant NRF-2021R1A2C2010893 and in part by Institute of Information \& communications Technology Planning \& Evaluation (IITP) grant funded by the Korea government(MSIT) (No.RS-2022-00155915, Artificial Intelligence Convergence Innovation Human Resources Development (Inha University).

% \printbibliography
\bibliography{ref}
\end{document}